\documentclass[11pt]{article}

\usepackage[final]{acl}
\usepackage{times}
\usepackage{latexsym}
\usepackage[utf8]{inputenc}
\usepackage[T1]{fontenc}
\usepackage{booktabs}
\usepackage{amsmath}
\usepackage{amssymb}
\usepackage{microtype}
\usepackage{xcolor}
\usepackage{colortbl}
\usepackage{graphicx}
\graphicspath{{figures/}}
\usepackage{tikz}
\usetikzlibrary{positioning, fit}
\usepackage{CJKutf8}
\usepackage{enumitem}
\usepackage{float}
\usepackage{natbib}

\title{One Head Away from Language Confusion: Single Attention Heads Across Model Families}
\author{
\begin{minipage}{0.98\textwidth}
\centering
{\bfseries\normalsize
Jiwon Jeong\textsuperscript{1}\quad Anjelo Jann Laroza\textsuperscript{2,*}\quad Chaewoon Yoon\textsuperscript{3,*}\\
Arjun Pillai\textsuperscript{4}\quad Pranav Kokati\textsuperscript{5}\quad Christian Hoang\textsuperscript{6}\quad Seunghyeok Hong\textsuperscript{7,$\dagger$}\\[0.35em]}
\vspace{0.35em}
{\normalfont\small
$^{1}$ Lotte Innovate \quad $^{2}$ Mapúa University \quad $^{3}$ University of Wisconsin--Madison\\
$^{4}$ Irvington High School \quad $^{5}$ Interlake High School \quad $^{6}$ FPT University\\
$^{7}$ Division of Social Science \& AI, Hankuk University of Foreign Studies\\
}
{\normalfont\scriptsize
107, Imun-ro, Dongdaemun-gu, Seoul, South Korea\\
Jiwon Jeong: \href{mailto:jiwon-jeong@lotte.net}{\texttt{jiwon-jeong@lotte.net}}; Anjelo Jann Laroza: \href{mailto:ajalaroza@mymail.mapua.edu.ph}{\texttt{ajalaroza@mymail.mapua.edu.ph}}\\
Arjun Pillai: \href{mailto:arjun.pillai360@gmail.com}{\texttt{arjun.pillai360@gmail.com}}; Seunghyeok Hong: \href{mailto:shongdr@gmail.com}{\texttt{shongdr@gmail.com}}
}
\end{minipage}
}
\hypersetup{
    pdftitle={One Head Away from Language Confusion: Single Attention Heads Across Model Families},
    pdfauthor={Jiwon Jeong, Anjelo Jann Laroza, Chaewoon Yoon, Arjun Pillai, Pranav Kokati, Christian Hoang, Seunghyeok Hong}
}
\newcommand{\teamtodo}[1]{}
\begin{document}

\maketitle
\begingroup
\renewcommand{\thefootnote}{\fnsymbol{footnote}}
\footnotetext[1]{Anjelo Jann Laroza and Chaewoon Yoon contributed equally.}
\footnotetext[2]{Corresponding author: \texttt{shongdr@gmail.com}.}
\endgroup

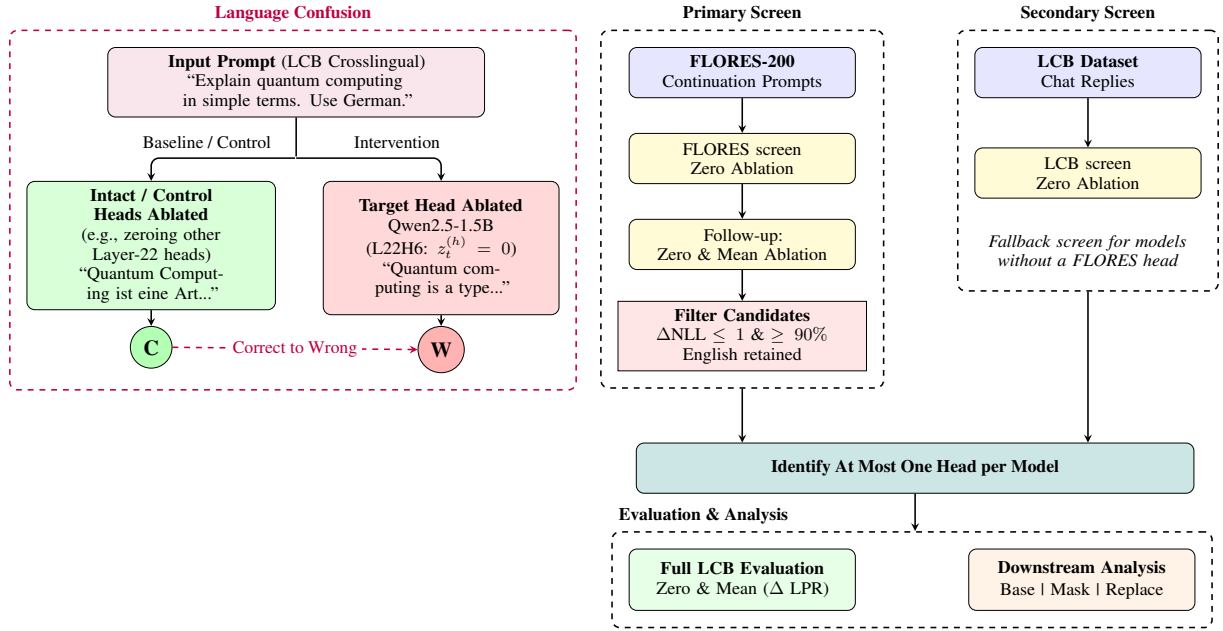
\begin{figure*}[t]
\centering
\resizebox{\linewidth}{!}{%
\begin{tikzpicture}[
    box/.style={rectangle, draw, rounded corners, align=center, minimum width=3.8cm, minimum height=0.9cm, font=\small},
    filterbox/.style={rectangle, draw, align=center, minimum width=3.8cm, minimum height=0.9cm, font=\small, fill=red!10, text width=4.2cm},
    group/.style={rectangle, draw, dashed, thick, rounded corners, inner sep=0.3cm},
    arrow/.style={-stealth, thick}
]
\def\primX{8.0}
\def\secX{14.2}
\def\midX{11.1} 
\node[box, fill=purple!10, text width=6.5cm, anchor=north] (prompt) at (0, 0) {\textbf{Input Prompt} (LCB Crosslingual)\\``Explain quantum computing in simple terms. Use German.''};
\node[box, fill=green!15, text width=4.2cm, anchor=north] (intact) at (-2.6, -2.4) {\textbf{Intact / Control}\\\textbf{Heads Ablated}\\(e.g., zeroing other Layer-22 heads)\\``Quantum Computing ist eine Art...''};

\node[box, fill=red!15, text width=113pt, anchor=north, minimum height=68pt] (ablated) at (2.6, -2.4) {\textbf{Target Head Ablated}\\Qwen2.5-1.5B (L22H6: $z_t^{(h)} = 0$)\\``Quantum computing is a type...''};
\node[circle, draw, fill=green!30, font=\bfseries, anchor=north] (C) at (-2.6, -5.0) {C};
\node[circle, draw, fill=red!30, font=\bfseries, anchor=north] (W) at (2.6, -5.0) {W};
\draw[thick] (prompt.south) -- (0, -1.96);
\draw[arrow, rounded corners] (-0.02, -1.99) -| (intact.north);
\draw[arrow, rounded corners] (0, -1.99) -| (ablated.north);
\node[above, font=\footnotesize, fill=white, inner sep=2pt] at (-1.6, -1.89) {Baseline / Control};
\node[above, font=\footnotesize, fill=white, inner sep=2pt] at (1.81, -1.89) {Intervention};
\draw[arrow] (intact.south) -- (C.north);
\draw[arrow] (ablated.south) -- (W.north);
\draw[arrow, dashed, thick, purple] (C.east) -- (W.west) node[midway, fill=white, font=\footnotesize, align=center, inner sep=2pt] {Correct to Wrong};
\node[group, fit=(prompt) (intact) (ablated) (C) (W), draw=purple, label={[font=\small, text=purple]above:\textbf{Language Confusion}}] (lc_group) {};
\node[box, fill=blue!10, anchor=north, text width=3.8cm] (flores_in) at (\primX, 0) {\textbf{FLORES-200}\\Continuation Prompts};
\node[box, fill=yellow!20, anchor=north, text width=3.8cm] (flores_screen) at (8, -1.54) {FLORES screen\\Zero Ablation};
\node[box, fill=yellow!20, anchor=north, text width=3.8cm] (flores_follow) at (8, -3.08) {Follow-up:\\Zero \& Mean Ablation};
\node[filterbox, anchor=north] (filter) at (8, -4.55) {\textbf{Filter Candidates}\\$\Delta$NLL $\le 1$ \& $\ge 90\%$ English retained};
\node[group, fit=(flores_in) (flores_screen) (flores_follow) (filter), label={[font=\small]above:\textbf{Primary Screen}}] (flores_group) {};
\node[box, fill=blue!10, anchor=north, text width=3.8cm] (lcb_screen_in) at (\secX, 0) {\textbf{LCB Dataset}\\Chat Replies};
\node[box, fill=yellow!20, anchor=north, text width=3.8cm] (lcb_screen) at (\secX, -1.8) {LCB screen\\Zero Ablation};
\node[draw=none, fill=none, font=\footnotesize\itshape, text width=3.8cm, align=center, anchor=north] (lcb_note) at (14.2, -3.23) {Fallback screen for models\\without a FLORES head};
\node[group, fit=(lcb_screen_in) (lcb_screen) (lcb_note), label={[font=\small]above:\textbf{Secondary Screen}}] (lcb_group) {};
\draw[arrow] (flores_in) -- (flores_screen);
\draw[arrow] (flores_screen) -- (flores_follow);
\draw[arrow] (flores_follow) -- (filter);
\draw[arrow] (lcb_screen_in) -- (lcb_screen);
\node[box, fill=teal!20, minimum width=10cm, anchor=north] (heads) at (11.1, -7.09) {\textbf{Identify At Most One Head per Model}};
\draw[arrow] (flores_group.south) -- (flores_group.south |- heads.north);
\draw[arrow] (lcb_group.south) -- (lcb_group.south |- heads.north);
\node[box, fill=green!10, minimum height=1.1cm, anchor=north, text width=3.8cm] (lcb_eval) at (8, -8.99) {\textbf{Full LCB Evaluation}\\Zero \& Mean ($\Delta$ LPR)};
\node[box, fill=orange!10, minimum height=1.1cm, anchor=north, text width=3.8cm] (downstream) at (14.09, -8.99) {\textbf{Downstream Analysis}\\ \vspace{0.05cm} \footnotesize Base | Mask | Replace};
\node[group, fit=(lcb_eval) (downstream), label={[font=\small, anchor=south west]north west:\textbf{Evaluation \& Analysis}}] (eval_group) {};
\draw[arrow] (heads.south) -- (heads.south |- eval_group.north);
\end{tikzpicture}%
}
\caption{Methodology for identifying and analyzing at most one head per instruction-tuned model, accompanied by a real-world demonstration of the correct-to-wrong language confusion metric.}
\label{fig:methodology}
\end{figure*}
\begin{abstract}
Large language models sometimes reply in an unexpected language, a failure known as language confusion.\ We ablate individual attention heads in 12 model checkpoints spanning seven model series, using FLORES-200 continuations in five languages and Language Confusion Benchmark (LCB) chat replies. In ten models, zeroing one head causes substantially more departures from the expected language than zeroing any other head in its layer; its LCB pass-rate drop persists, though reduced, under mean ablation. Seven heads move 8.0--51.5\% of all FLORES continuations out of the prompt's language, versus at most 1.4\% for other same-layer heads; the other three mainly affect explicitly requested languages. These seven heads matter less in base models given plain text. In two models, replacing the values a head reads at the requested-language name with another language's mean makes replies fail and shifts some into that language, unlike a random change of equal size. Replacing a head's output with another language's mean redirects 84--99\% of scored non-English monolingual LCB replies into that language in five models. Yet the requested language's mean improves crosslingual replies beyond every control head only in the two Gemma models, whose baselines are low. Code and results are available in \href{https://github.com/christian-hoang-04/anonymous-release}{the public repository}.
\end{abstract}

\section{Introduction}

Large language models (LLMs) are expected to reply in the language of the prompt
or in a language explicitly requested by the user, yet they sometimes reply in
another language. \citet{marchisio-etal-2024-understanding} call this failure
\emph{language confusion} and measure it with the Language Confusion Benchmark
(LCB). In their evaluation, base and English-centric instruction-tuned models
are more prone to it, and even the strongest models do not always reply in the
expected language. What happens inside the model when a reply leaves the
expected language is less well understood.

Mechanistic studies suggest that output language information is partly separable from content and that later layers contribute to language-specific
decoding \citep{wendler-etal-2024-llamas,dumas2025separating,harrasse2026tracing}.
Prior work has also shown that output language can be steered or repaired
through small sets of neurons
\citep{tang-etal-2024-language,nie2025mechanistic} or attention heads
\citep{liu2025lahis}, including heads involved in target language identification
during few-shot translation in base models \citep{lasnier2026translation}.
It remains unclear whether language adherence in chat replies
can be disproportionately sensitive to a single head, whether similar
dependencies appear across model families, and whether heads identified
on text continuations also influence the language of chat replies. A large language adherence effect need not
coincide with a large increase in language modeling loss: zeroing Qwen2.5-7B's
L19H1 raises the loss by only 0.026 nats but moves 20\% of FLORES
continuations out of the prompt's language.

We address these questions by ablating attention heads one at a time in 12
models from seven model series, using
FLORES-200 continuations in five languages and monolingual and crosslingual
chat prompts from LCB (Figure~\ref{fig:methodology}). We distinguish
correct-to-wrong from wrong-to-correct language changes, set aside heads whose
removal sharply raises language-modeling loss or frequently moves English
continuations out of English, and compare each selected head with the other
heads in its layer. For models without a qualifying head on FLORES, we screen
all heads on LCB. We then compare the seven FLORES heads with their base
checkpoints, examine what two heads read, and replace head outputs with
language-conditioned means. Our main findings are:

\begin{enumerate}[label=(\arabic*),leftmargin=*,nosep]
\item In ten of the twelve models, zeroing one head causes far more departures
from the expected language than zeroing any other head in its layer. Its effect
on LCB language adherence persists, though reduced, under mean ablation. Seven
of these heads move 8.0--51.5\% of all FLORES continuations out of the prompt's
language; the other three mainly affect explicitly requested languages.

\item The seven FLORES heads have smaller effects in the base models given
plain text; in Qwen2.5-1.5B, the difference appears mainly with the chat
template.

\item In Qwen2.5-1.5B and Gemma-3-1B, replacing the values a head reads at
tokens naming the requested language with mean values from prompts naming
another language makes replies fail and shifts some into that language, which a
random change of the same size does not do.
Masking the head's attention to these tokens also reproduces part of the
effect of removing the head.

\item Replacing a head's output with its mean for another language puts
84--99\% of scored replies to non-English monolingual LCB prompts entirely
in the replacement language in five models. However, the requested
language's mean improves crosslingual replies beyond every control head
only in the two Gemma models.
\end{enumerate}

\section{Related Work}

\paragraph{Head ablation.}
Mechanistic analyses study how internal components contribute to model
behavior \citep{elhage2021} and localize some behaviors in small circuits
of attention heads \citep{wang2022interpretability,conmy2023towards}.
Many heads can be removed with little performance loss, sometimes
reducing layers to a single head \citep{michel2019sixteen}, while
specialized heads in machine translation tend to survive pruning
\citep{voita-etal-2019-analyzing}. Some capabilities, however, depend
on small subsets of heads: retrieval heads constitute fewer than 5\%
of the heads in the models studied, yet their ablation substantially
impairs long-context retrieval, unlike ablating random non-retrieval
heads \citep{wu2024retrieval}. We investigate the effects of
single-head ablation on output-language behavior rather than
aggregate task performance.

\paragraph{Multilingual representations.}
In Llama-2, analyses of intermediate layers suggest a concept space
closer to English than to other languages
\citep{wendler-etal-2024-llamas}, while activation patching shows that
the output language is encoded at an earlier layer than the concept
\citep{dumas2025separating}.
Language-specific decoding emerges in later layers
\citep{harrasse2026tracing}, and cross-lingual factual
inconsistencies have been linked to failures in the late transition
to language-specific representations \citep{wang2025lost}.
Key language-related neurons also concentrate in the first and last
layers \citep{zeng2025converging}. These findings motivate screening
attention heads across all layers rather than at a preselected depth.

\paragraph{Language-selective neurons.}
\citet{tang-etal-2024-language} identify language-specific neurons,
mostly in the first and last layers, and steer output languages by
activating or deactivating them.
\citet{gurgurov2025language} study the identification and manipulation
of such neurons across 21 languages.
\citet{tezuka2025transfer} propose that specific MLP neurons mediate
transitions between language-specific latent spaces and a shared
semantic space. Among neuron-level studies,
\citet{nie2025mechanistic} trace language confusion in English-centric
models to transition failures in the final layers and mitigate them
by editing neurons identified through comparison with a
multilingual-tuned counterpart. We examine whether individual
attention heads can have similarly large causal effects on language adherence across model families.

\paragraph{Attention heads and language control.}
LAHIS \citep{liu2025lahis} identifies language-specific and
language-general heads in Aya-23-8B, Llama-3.2-3B, and Mistral-7B-v0.1,
using language-specific heads to reduce off-target generation.
\citet{lasnier2026translation} study few-shot translation in base
models without explicit instructions. Through activation patching,
they identify distinct sparse head sets for target-language
identification and meaning preservation. Steering these heads enables
instruction-free translation, while their ablation selectively
disrupts the corresponding functions.

Their language heads provide a close precedent for ours.
In their released base-model scores \citep{mitra2026repo}, averaged
over 20 translation directions, the heads we identify in
instruction-tuned Qwen3-1.7B and Llama-3.2-1B each rank first within
their models, while the Gemma-3-1B head ranks third
(Appendix~\ref{app:mitra}).
Because the scores come from base models, this agreement does not
establish identical functions.

Building on these studies, we examine whether keeping chat replies in the
expected language is disproportionately sensitive to individual heads
across model families, both when that language is implicit and when it is
explicitly requested. We further compare these dependencies with
base checkpoints, investigate what selected heads read, and test
whether replacing their outputs can restore the requested language.

\paragraph{Language confusion and inference-time steering.}
LCB \citep{marchisio-etal-2024-understanding} evaluates language
confusion in monolingual and crosslingual prompts, including
explicit requests for a reply language; we use its prompts and
line-level pass rate.
\citet{oh2026ola} find failures in inferring the expected reply
language from code-switched prompts, including biases toward
non-English replies and within-response language switching.
Language-specific steering vectors extracted from a multi-parallel
corpus can reduce language confusion while preserving task
performance \citep{sterz2025recover}.
Function vectors add task-conditioned mean outputs of several
heads to the residual stream to induce a task
\citep{todd2024function}. We instead replace one head's output
with its mean for a language to probe its role in output-language
control, rather than to propose a general mitigation method.

\section{Setup}
\label{sec:setup}

\subsection{Models and Data}
\label{sec:models_data}

We evaluate 12 model checkpoints
across seven model series: Qwen2.5 1.5B, 3B and 7B \citep{qwen2025}, Qwen3 1.7B
and 4B \citep{yang2025qwen3}, Gemma 3 1B and 4B \citep{gemmateam2025gemma3},
OLMo 2 1B \citep{olmo2024olmo2}, OLMo 3 7B \citep{olmo2025olmo3}, Llama 3.2 1B
and 3B \citep{grattafiori2024llama3}, and SmolLM3 3B \citep{bakouch2025smollm3}.

For implicit continuation, we use 2,500 prompts from FLORES-200
\citep{nllbteam2022noLanguageLeftBehind}, 500 each in English, French, German,
Spanish and Italian: the first 495 aligned devtest sentences that pass a length
and letter-ratio filter, and five hand-written sentence starts per language
(Appendix~\ref{app:protocol}). The five starts and the first 20 sentences per
language form the screening set.

For chat replies, we use the Language Confusion Benchmark
(LCB; \citealp{marchisio-etal-2024-understanding}). The five-language
evaluation includes monolingual prompts written in French, German, Spanish, or
Italian (800 prompts), plus 200 English prompts as controls. Crosslingual
prompts are written in English and request a reply in one of the four
non-English languages (1,196 prompts). Instruction-tuned models receive every
prompt as one user turn in their own chat template; base checkpoints receive
FLORES prompts as plain text (templates, precision and decoding in
Appendix~\ref{app:protocol}).

\subsection{Interventions}
\label{sec:interventions}

Let $z^{(h)}_t$ be the output slice of attention head $h$ at sequence position
$t$, immediately before the attention output projection $o_{\mathrm{proj}}$,
which mixes the heads' contributions across all hidden dimensions. Our primary
intervention sets this slice to zero at every position, while the prompt is
processed and while the reply is generated; Section~\ref{sec:reply_language_mechanism}
also intervenes in one phase at a time.

Since interpretability results can depend on how an activation is replaced
\citep{zhang2024best}, we also report mean ablation, which replaces
$z^{(h)}_t$ with its mean $\mu_h$ in the unmodified model over the prompt
positions of the 2,500 FLORES prompts, chat-template tokens included; the same
mean is used for FLORES and LCB. For language-specific replacement, we instead average
the head's output over the FLORES prompts in language $\ell$ and the unmodified
model's greedy continuations of them, from the first token of the user's text to
the end of the continuation, giving $\mu_{h,\ell}$; Section~\ref{sec:output_carries}
replaces the head's output with the mean for a chosen language. Unlike $\mu_h$,
this mean also covers generated tokens; the prompt language's own mean, computed in
the same way, serves as its control (Section~\ref{sec:output_carries},
Appendix~\ref{app:zero_mean_diagnostics}).

\subsection{Measuring Language Confusion}
\label{sec:metrics}

On FLORES, each model greedily generates up to 40 new tokens, and
\texttt{langdetect} \citep{shuyo2010} assigns the language $\mathcal{L}(y)$ of
each continuation $y$. Let $\ell_j$ be the language of prompt $j$, and let $y_j$
and $y_j^{-h}$ be the baseline and head-ablated continuations. The directional
correct-to-wrong rate is
\begin{equation}
\begin{split}
C\to W(h) = \frac{1}{n}\sum_{j=1}^{n}
  \mathbf{1}\bigl[\,&\mathcal{L}(y_j)=\ell_j\\
  &{}\wedge\mathcal{L}(y_j^{-h})\ne\ell_j\bigr],
\end{split}
\label{eq:c2w}
\end{equation}
where $n$ is the number of prompts in all five languages (2,500, or 125 in the
screen). Unlike a raw switch rate, $C\to W$ ignores changes away from an
already wrong label; the reverse change, $W\to C$, is at most 0.002 for each of
the seven heads found on FLORES.

On LCB, replies are generated greedily for up to 100 new tokens and scored with
the benchmark's rules \citep{joulin2016fasttext,marchisio-etal-2024-understanding}:
fastText labels each line of at least five words, a reply passes only when all
its scored lines are in the expected language, and replies with no such line are
not scored. The line-level pass rate (LPR) is the pass rate averaged over the
benchmark's sources. For each head, $\Delta\mathrm{LPR}$ is the paired mean
difference in pass outcomes between ablated and unablated replies over the
prompts scored in both (Appendix~\ref{app:protocol}), computed separately for monolingual and crosslingual prompts; since it weights prompts equally, it can differ from the difference of
the source-averaged LPRs. A negative value means less adherence to the expected
language. Its 95\% confidence interval comes from 2,000 paired bootstrap
resamples of the prompts.

\begin{table*}[t]
\centering\small
\setlength{\tabcolsep}{3.5pt}
\begin{tabular}{@{}lclccccccccc@{}}
\toprule
 & & & \multicolumn{4}{c}{FLORES $C\to W$} & & \multicolumn{2}{c}{LCB monolingual} & \multicolumn{2}{c}{LCB crosslingual} \\
\cmidrule(lr){4-7}\cmidrule(lr){9-10}\cmidrule(lr){11-12}
Model & H & Head & Zero & Mean & Others & Base & $\Delta$NLL & LPR & $\Delta$LPR & LPR & $\Delta$LPR \\
\midrule
\rowcolor{gray!15}\multicolumn{12}{c}{Heads found on FLORES} \\
Qwen2.5-1.5B & 12 & L22H6 & 0.500 & 0.450 & 0.007 & 0.150 & $+$0.228 & 0.982 & $-$0.273 & 0.704 & $-$0.276 \\
Qwen2.5-3B & 16 & L27H13 & 0.515 & 0.338 & 0.014 & 0.202 & $+$0.234 & 0.982 & $-$0.537 & 0.888 & $-$0.449 \\
Qwen2.5-7B & 28 & L19H1 & 0.200 & 0.018 & 0.006 & 0.025 & $+$0.026 & 0.984 & \textcolor{gray}{$-$0.009} & 0.950 & $-$0.074 \\
Qwen3-1.7B & 16 & L18H12 & 0.324 & 0.028 & 0.002 & 0.067 & $+$0.128 & 0.985 & $-$0.149 & 0.823 & $-$0.508 \\
Gemma-3-1B & 4 & L11H3 & 0.413 & 0.089 & 0.002 & 0.001 & $+$0.720 & 0.984 & $-$0.628 & 0.118 & $-$0.091 \\
Gemma-3-4B & 8 & L24H0 & 0.214 & 0.012 & 0.000 & 0.148 & $+$0.202 & 0.990 & $-$0.259 & 0.133 & $-$0.107 \\
OLMo-2-1B & 16 & L12H8 & 0.080 & 0.056 & 0.005 & 0.005 & $+$0.136 & 0.986 & $-$0.272 & 0.931 & $-$0.335 \\
\rowcolor{gray!15}\multicolumn{12}{c}{Heads found on LCB} \\
Qwen3-4B & 32 & L24H27 & 0.016$^{s}$ & -- & 0.008$^{s}$ & -- & $+$0.065 & 0.982 & \textcolor{gray}{$+$0.000} & 0.893 & $-$0.076 \\
Llama-3.2-1B & 32 & L8H25 & 0.048$^{s}$ & -- & 0.016$^{s}$ & -- & $+$0.067 & 0.997 & $-$0.016 & 0.874 & $-$0.711 \\
Llama-3.2-3B & 24 & L13H19 & 0.000$^{s}$ & -- & 0.000$^{s}$ & -- & $+$0.008 & 0.993 & \textcolor{gray}{$+$0.000} & 0.911 & $-$0.335 \\
\rowcolor{gray!15}\multicolumn{12}{c}{Not counted (Section~\ref{sec:one_head_flores})} \\
OLMo-3-7B & 32 & L14H25 & 0.134 & 0.007 & 0.010 & 0.104$^{s}$ & $+$0.202 & 0.972 & $-$0.041 & 0.874 & $-$0.018 \\
SmolLM3-3B & 16 & \multicolumn{10}{l}{none in either screen (L1H12 breaks generation)} \\
\bottomrule
\end{tabular}
\caption{Head-ablation effects across the 12 evaluated models.
H: heads per layer. FLORES: Zero and Mean are $C\to W$ (Eq.~\eqref{eq:c2w})
under zero and mean ablation; Others is the largest same-layer Zero, and Base is
Zero in the base checkpoint. $\Delta$NLL: loss change under zero ablation. LCB:
baseline pass rate (LPR) on the non-English prompts and its paired change under
zero ablation ($\Delta$LPR; Section~\ref{sec:metrics}). $^{s}$: 125-prompt FLORES
screen; gray: 95\% CI includes zero; --: not evaluated. Heads found on LCB are
compared on LCB (Section~\ref{sec:one_head_lcb}); all screens are in
Appendix~\ref{app:head_screens}. For Qwen3-4B, Table~\ref{tab:heads} reports the all-prompt estimate ($-0.076$; 1,196 prompts, 1,187 pairs), while Section~\ref{sec:one_head_lcb} reports the held-out estimate ($-0.073$; 1,096 prompts, 1,089 pairs), excluding the 100-prompt selection screen.}
\label{tab:heads}
\end{table*}

\subsection{Choosing the Head}
\label{sec:head_selection}

All 12 models are screened on FLORES; eight have a head that qualifies there and
go to a 2,500-prompt follow-up, and the other four are screened on LCB
(Figure~\ref{fig:methodology}). On the 125 screening prompts we ablate every head
and apply two conditions: $\Delta$NLL $\leq 1$, where $\Delta$NLL is the change
in mean per-token loss (in nats) on 100 FLORES dev sentences per language, given
as plain text and averaged over the five languages, and at least 90\% of English
continuations kept in English, meant to exclude heads whose removal disrupts
generation in general or on English prompts. This $\Delta$NLL $\leq 1$ cutoff is the primary loss guard for screening and head selection. The stricter $\Delta$NLL $\leq 0.1$ criterion in Appendix~\ref{app:head_screens} selects one additional low-loss follow-up candidate; it is not the primary eligibility cutoff. A model goes to the follow-up when
a head that meets both conditions reaches a $C\to W$ of 0.10 on the screen. The
follow-up reruns all heads of the layers that hold the strongest screened heads
(Appendix~\ref{app:head_screens}), and we report the one with the largest
$C\to W$ on all 2,500 prompts among those that meet both conditions, with English
retention now measured on all 500 English prompts. We compare each selected head
with the other heads of its layer rather than across architectures, and in each
full LCB evaluation we also ablate three randomly chosen heads of the same layer
as controls (one for Llama-3.2-1B). Because the head is chosen on all 2,500
prompts, its FLORES effect is conditioned on that choice. The head itself is
chosen without LCB data, so LCB tests it on a separate task.
LCB did, however, decide which FLORES heads are counted: OLMo-3-7B also
qualifies on FLORES, but we report it separately, by a criterion set after its
LCB results that all ten counted heads meet (Section~\ref{sec:one_head_flores}).

The LCB screen ablates every head on 100 crosslingual LCB prompts (25 per
requested language). Among the heads that meet both conditions on the FLORES
screen, the one with the largest drop in the pooled pass rate goes to the full
LCB evaluation, provided the drop is at least 0.10; for these heads, the
comparison with the other heads of the layer comes from this screen, and
Section~\ref{sec:one_head_lcb} reports their effect on the crosslingual prompts
not used to select them. Section~\ref{sec:specificity} reruns this screen on the
models with a FLORES follow-up as a check.

\section{A Single Head Stands Out in Ten Models}
\label{sec:one_head}

We count ten heads (Table~\ref{tab:heads}): seven found on FLORES and three found
on LCB that act mainly on explicitly requested languages. We report OLMo-3-7B's
FLORES head separately, and SmolLM3-3B has no qualifying head.

\subsection{Passage Continuations (FLORES)}
\label{sec:one_head_flores}

In seven models, zeroing one head moves 8.0--51.5\% of all FLORES
continuations out of the prompt's language, versus at most 1.4\% for any other
head in its layer. Most departures go to English in Qwen2.5-1.5B, Qwen3-1.7B and
Gemma-3-1B, whereas 441 of the 500 departures in Qwen2.5-7B go to Chinese
(Appendix~\ref{app:checks}).

We report OLMo-3-7B's L14H25 (13.4\%) separately: its LCB effects are small
($-$0.041 and $-$0.018) and vanish under mean ablation ($+$0.001 and $+$0.015),
whereas each of the ten counted heads still lowers at least one LCB pass rate
under mean ablation (Section~\ref{sec:one_head_lcb}).

\subsection{Chat Replies (LCB)}
\label{sec:one_head_lcb}

Each of the ten heads lowers at least one LCB pass rate, with the 95\% CI of its
paired change below zero (Table~\ref{tab:heads}, Appendix~\ref{app:lcb}); the
upper end of that interval is also below the change of every randomly chosen
same-layer control head. For the seven FLORES heads, the effects are roughly
balanced across monolingual and crosslingual prompts in Qwen2.5-1.5B, Qwen2.5-3B and OLMo-2-1B, mostly crosslingual in Qwen3-1.7B
and Qwen2.5-7B, and mostly monolingual in the two Gemma models, whose
crosslingual baselines are low.  In three of the four
models screened on LCB,
the screen selects L8H25 (Llama-3.2-1B), L13H19 (Llama-3.2-3B) and L24H27
(Qwen3-4B). Their ablation turns 71.0, 32.3 and 11.2\% of screening replies from
pass to fail, against at most 4.0\% for any other head of their layer, and lowers
the pass rate on the crosslingual prompts outside the screen by 0.711, 0.336 and
0.073 (Appendix~\ref{app:lcb}), with little monolingual change.

Under mean ablation, each of the ten heads still lowers the pass rate of the
task it affects most, keeping 39--89\% of the zero-ablation drop with the 95\%
CI below zero (Appendix~\ref{app:lcb}). Across 14 target languages, the selected
head reduces crosslingual pass rates in every language in five of six tested
models, and in all but Japanese and Russian in Qwen2.5-1.5B
(Appendix~\ref{app:lcb14}).

\subsection{Specificity Checks}
\label{sec:specificity}

\paragraph{General degradation.}
On plain text, the seven FLORES heads raise the loss by 0.026 to 0.720 nats,
against a layer mean of at most 0.022 for the other heads, so their removal is
not free. All seven keep every English continuation in English, beyond the
90\% the selection requires. The ten heads also leave English LCB prompts,
which play no part in selection, at pass rates of 0.990--1.000. In five models,
continuations switching to English stay closer to the prompt in embedding space
(0.558--0.683) than the next sentence of the same article (0.384;
Appendix~\ref{app:checks}). These checks argue against broad generation failure
but do not establish that output quality is preserved.

\paragraph{Head specificity and detectors.}
Random same-layer heads change LCB pass rates by at most 0.038. Replacing the
head's output with a random vector of the same norm moves at least as many
continuations as zeroing it (Appendix~\ref{app:zero_mean_diagnostics}), so $C\to W$ alone does
not show that the head's output carries language-specific information; in the
four models where another head of the layer has a larger output norm, zeroing
that head moves at most 1.4\%. Relabeling with langid, fastText or a two-of-three
vote changes FLORES $C\to W$ by at most 0.02 for all seven heads.

\paragraph{Found again on LCB alone.}
The same LCB screen, run on the eight models with a FLORES follow-up (with
monolingual prompts for the two Gemma models, whose crosslingual baselines of
0.118 and 0.133 leave little room for a drop), ranks the FLORES head first among
all eligible heads in five of the seven counted ones (Appendix~\ref{app:lcb}). In Qwen2.5-1.5B it ranks second, behind
L16H9, which reduces crosslingual pass rates by 0.459 but leaves monolingual
replies unchanged; in Qwen2.5-7B it ties for fourth, behind L0H1, whose effect
vanishes under mean ablation.

\section{Base Models and Post-Training Checkpoints}
\label{sec:base_instruct}

Do these heads already matter before post-training? We zero the seven FLORES
heads in the corresponding base models on the same 2,500 prompts, given as plain
text (Base in Table~\ref{tab:heads}). Every head has a smaller effect there:
0.001 to 0.202, or 0.2--69\% of the instruction-tuned value. The base models keep
the prompt language in 93--99\% of unablated continuations, and counting only
those continuations changes their values by at most 0.003
(Appendix~\ref{app:checkpoint_details}), but the comparison is confounded by
input format, since only the instruction-tuned
models receive a chat template.

\begin{figure}[t]
\centering
\includegraphics[width=\columnwidth]{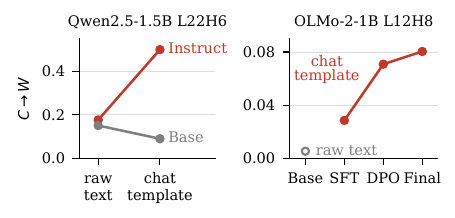}
\caption{FLORES $C\to W$ on 2,500 prompts when the selected head is zeroed.
Left: Qwen2.5-1.5B L22H6 in the Base and Instruct checkpoints, on raw text and
with the Instruct chat template. Right: OLMo-2-1B L12H8 in the released
checkpoints (Base on raw text, the others with the chat template). The y-axis
ranges differ.}
\label{fig:base}
\end{figure}

\paragraph{Input format in Qwen2.5-1.5B.}
Crossing checkpoint and input format (Figure~\ref{fig:base}, left), zeroing
L22H6 moves 17.6\% (Instruct) and 15.0\% (Base) of continuations out of the
prompt's language on raw text, but 50.0\% (Instruct) and 8.9\% (Base) with the
Instruct chat template. The head already matters before post-training, and its
effect is largest when the Instruct checkpoint receives its own template,
suggesting an interaction between post-training and input format.

\paragraph{OLMo-2 checkpoints.}
Across the released OLMo-2-1B checkpoints (Figure~\ref{fig:base}, right), the
FLORES effect of L12H8 grows from 0.005 (base) to 0.028 (SFT), 0.071 (DPO) and
0.080 (final), while its crosslingual LCB effect is already $-$0.247 after SFT
(Appendix~\ref{app:checkpoint_details}). The SFT, DPO and final checkpoints share
a chat template, so the later increases are not due to format alone, but the
base-to-SFT step remains confounded by format, and the final checkpoint adds
reinforcement learning with verifiable rewards after DPO, so the checkpoints show
how the dependence changes along the released training sequence without
isolating its cause.

\section{What the Head Does}
\label{sec:what_head_does}

\subsection{Timing, Attention and Values}
\label{sec:reply_language_mechanism}

\paragraph{When the head matters.}
We first test Qwen2.5-1.5B L22H6 on 96 monolingual LCB prompts (24 per
language), sampled independently of earlier ablation results; none names a reply
language, and all pass at baseline. Zeroing the head only in the decoding steps
after the first generated token makes 31 replies fail, against 2 when it is
zeroed only during prompt processing and 35 when it is zeroed in both; zeroing
L22H8, another head of the layer, in both makes none fail. The head thus matters mainly while the reply is
generated. All 24 Italian replies fail when it is zeroed during decoding, as in
the full LCB run (Italian monolingual LPR 1.00 to 0.00).

\begin{table}[t]
\centering\footnotesize
\setlength{\tabcolsep}{3pt}
\begin{tabular}{@{}lrrrr@{}}
\toprule
 & \multicolumn{2}{c}{Qwen2.5-1.5B} & \multicolumn{2}{c}{Gemma-3-1B} \\
\cmidrule(lr){2-3}\cmidrule(lr){4-5}
\multicolumn{5}{@{}l}{\textit{Attention from the last prompt token}} \\
To the name & \multicolumn{2}{r}{0.773} & \multicolumn{2}{r}{0.132} \\
To as many nearby tokens & \multicolumn{2}{r}{0.003} & \multicolumn{2}{r}{0.009} \\
Control head, to the name & \multicolumn{2}{r}{0.039} & \multicolumn{2}{r}{0.093} \\
\midrule
 & Fail & Other & Fail & Other \\
\cmidrule(lr){2-3}\cmidrule(lr){4-5}
Name masked & 34 & 0 & 32 & 0 \\
Nearby tokens masked & 0 & 0 & 0 & 0 \\
Control head's mask & 0 & 0 & 2 & 0 \\
Head zeroed & 105 & 0 & 36 & 0 \\
Same-language values & 1 & 0 & 0 & 0 \\
Other-language values & 90 & 44 & 43 & 24 \\
Random-direction values & 4--8 & 0 & 1--6 & 0 \\
Second-language values & 217 & 208 & 25 & 16 \\
L11H2's name masked & -- & -- & 2 & 0 \\
\bottomrule
\end{tabular}
\caption{Mechanism checks for Qwen2.5-1.5B L22H6 and Gemma-3-1B L11H3 on 480
crosslingual LCB prompts that name the requested language. The control head
attends most to the name among the other heads of the layer (L22H7, L11H1). Values: the
values the head reads at the name, replaced by a same-, other- or
second-language mean, or moved by the other-language distance in a random
direction (three seeds).
All interventions act in the decoding steps after the first token. Fail:
baseline-passing replies (323 and 51) that fail; Other: replies entirely in
German, or French for German prompts (in the second-language row, in that
language), of the 471 to 476 scored in each condition; --: Gemma-3-1B only.
Per-sample results in Appendix~\ref{app:mechanism}.}
\label{tab:attention}
\end{table}

\paragraph{Attention to the language name.}
We then test both heads on 480 crosslingual LCB prompts (120 per language) that
name the requested language in English, sampled without using any model's
replies: 96 first, then 384 more in a replication fixed before it was run
(Appendix~\ref{app:mechanism}). At the last prompt position, both heads attend
to the name more than any other head of their layer, and far more than to as
many nearby tokens (Table~\ref{tab:attention}); the runner-up (L22H7, L11H1,
chosen before any intervention) serves as the control head. Over the first eight
reply tokens, L22H6 still leads, whereas in Gemma-3-1B L11H2 attends slightly more
than L11H3 (0.079 against 0.064; Appendix~\ref{app:mechanism}). Masking the head's
attention to the name in the decoding steps after the first generated token
makes 34 of 323 baseline-passing Qwen2.5-1.5B replies and 32 of 51 Gemma-3-1B
replies fail, while masking its attention to the nearby tokens, or the control
head's attention to the name, makes at most two fail. Prespecified two-sided
exact McNemar tests \citep{mcnemar1947note} against each control give
$p=0.031$ and $p=0.016$ on the first 96 prompts (6 of 62 and 7 of 12 replies,
uncorrected) and $p<10^{-6}$ on the 384 new ones. Masking L11H2's attention to
the name makes only 2 of the 51 Gemma-3-1B replies fail ($p<10^{-8}$ against
L11H3; Appendix~\ref{app:mechanism}).

Zeroing the head in the same steps makes 105 and 36 replies fail, including 33
of the 34 and 24 of the 32 mask failures: masking reproduces part of the effect
of zeroing the head, and the 31 monolingual failures suggest that the head does
more than read the name. At most four baseline-failing replies pass under any of
these conditions.

\paragraph{What the name carries.}
The name mask also removes far more attention than either control, so it does
not separate the identity of the name from the amount of attention removed. We
therefore leave the head's attention weights as computed and replace only the
values it reads
at the name with their mean over other prompts that name the same language or
another language (German, or French for German prompts). The same-language values
make 1 and 0 baseline-passing replies fail, while the other-language values make
90 and 43 fail and put 44 and 24 of all scored replies entirely in the other
language, against none under the same-language values ($p<10^{-6}$ for each
comparison, exact McNemar tests fixed before the main runs). The other-language
values change the head's output more (mean norm of the change 2.0 against 0.4 in
Qwen2.5-1.5B, 0.36 against 0.10 in Gemma-3-1B), so we added two controls, with
their readings fixed before they were run (Appendix~\ref{app:mechanism}). Moving
each name value by the same distance in a random direction changes the output at
least as much (2.03 to 2.04 and 0.37) but makes only 4 to 8 and 1 to 6 replies
fail across three seeds ($p<10^{-10}$ against the other-language values) and moves
none into German or French. Values from a second language (Spanish, Italian or
French) put 208 and 16 replies entirely in that language, against none under the
same-language values ($p<10^{-4}$) and at most 8 when the head is zeroed. The values the head reads at the name
thus carry information about the language they name into the reply.

\subsection{Replacing the Output with Language Means}
\label{sec:output_carries}

We separate redirecting replies into another language from restoring the
language a user requested.

\begin{figure}[t]
\centering
\includegraphics[width=\columnwidth]{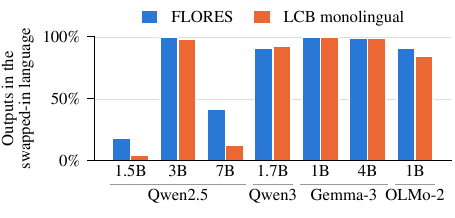}
\caption{Share of outputs in the swapped-in language when the head's output is
replaced by its FLORES mean for another language (German, or French on German
prompts). FLORES: the 2,000 non-English continuations, LCB monolingual: replies
to the non-English monolingual prompts that are entirely in that language.
Exact values in Table~\ref{tab:zero_mean_diagnostics}.}
\label{fig:swap}
\end{figure}

\paragraph{Swapping the language.}
In four models, mean ablation keeps only 6--22\% of the zero-ablation effect on
FLORES (Table~\ref{tab:heads}). Replacing the output at every position with
another language's mean $\mu_{h,\ell}$ (Section~\ref{sec:interventions}) instead
redirects the output language in five of the seven models
(Figure~\ref{fig:swap}): 90.4--99.7\% of the 2,000 non-English FLORES
continuations, on the prompts used to compute the means, are then in the
swapped-in language, while in these five the prompt language's own mean moves
at most 0.1\% out of it. With the same means, computed on FLORES only,
84.4--99.4\% of the replies to non-English monolingual LCB prompts are entirely
in that language, and swapping any of three same-layer control heads per model
puts none there. Qwen2.5-1.5B and Qwen2.5-7B show far lower rates (18.2\% and
41.7\% on FLORES, 4.3\% and 12.2\% on LCB).

\paragraph{Steering toward the requested language.}
We count replacing the output with the mean for the language a crosslingual LCB
prompt requests as an improvement when the lower end of the 95\% CI of the pass
rate change is above zero and above every control head's change. This holds
only in the two Gemma models, whose crosslingual baselines are low: from 0.118 to 0.560 in Gemma-3-1B (paired
$\Delta$LPR $+0.440$ [$+0.410$, $+0.469$]) and from 0.133 to 0.213 in
Gemma-3-4B ($+0.079$ [$+0.064$, $+0.095$]). The other five start at 0.70 to
0.95, and in Qwen2.5-1.5B, at 0.704, the pass rate falls by 0.020. Setting the head's output can thus redirect the reply language, but it does
not reliably restore a requested one (content of the corrected replies and
further diagnostics in Appendix~\ref{app:zero_mean_diagnostics}).

\section{Discussion and Conclusion}
\label{sec:discussion}

We ablated attention heads one at a time in 12 models from seven model series. In ten
of them, removing one head moves far more outputs out of the expected language
than removing any other head of its layer, and its LCB effect persists, though
reduced, under mean ablation. Which task the head affects varies: some heads
stand out on FLORES continuations, others mainly on explicitly requested
languages, and in Qwen2.5-1.5B and Qwen2.5-7B the LCB screen favors a different
head from the FLORES screen. As a secondary exploratory comparison, the dependence on the selected head is weaker in most larger models: its largest LCB drop is 0.28 to 0.71 in the seven counted 1--3B models, whereas in three of the four models with 4B or more parameters it is at most 0.076, a comparison that does not separate size from model family. In Qwen2.5-7B, the head the LCB screen prefers (L0H1) lowers crosslingual pass rates by 0.271 under zero ablation but not under mean ablation (Appendix~\ref{app:lcb}).

The seven FLORES heads matter less in the base models given plain text, and the
Qwen2.5-1.5B format runs and the OLMo-2-1B checkpoints suggest that this
dependence changes with post-training and prompt format. A head's importance, its
mechanism and its use for steering are related but distinct. In two models, the
values the head reads at the name of the requested language carry information
about that language into the reply, beyond the size of the change, yet the name
is not its only input. Replacing the head's output with another language's mean
redirects the output language in five models, but Qwen2.5-1.5B L22H6, with one
of the largest ablation effects, redirects only 18.2\% of FLORES continuations,
and the requested language's mean meets our criterion for improving crosslingual
replies only in the two Gemma models. A head can thus be important for keeping
the reply language without being a reliable way to set it. 

% \section{Conclusion}
% \label{sec:conclusion}

% We ablated attention heads one at a time in 12 instruction-tuned models to ask
% whether keeping replies in the expected language rests on single heads. In ten
% of them, removing one head moves far more outputs out of the expected language
% than removing any other head of its layer and clearly lowers LCB pass rates,
% and the seven heads found on FLORES matter less in the base models. Replacing
% the output of these seven heads with another language's mean redirects the
% output language in five of them, but replacing it with the requested language's
% mean meets our criterion for improving crosslingual replies only in the two
% Gemma models, both with low baselines. Language retention can thus depend
% disproportionately on a single head, but this dependence varies with the model,
% the task and the checkpoint, and a head's importance, its mechanism and its use
% for steering are related but distinct properties. Future work could test
% whether such heads appear in larger models and more languages, and which
% training stage creates the dependence.

\section*{Limitations}

\paragraph{Evaluation coverage.}
FLORES covers five European languages, and the main LCB evaluation four
requested non-English languages plus English controls; the 14-language
extension covers six of the twelve models. The results therefore do not
establish generality across languages, scripts or multilingual settings. LCB's
line-level metric leaves replies without a line of at least five words
unscored, which can omit short answers and some script switches where words are
not separated by spaces. LPR measures language adherence, not whether a reply is
correct.

\paragraph{Scope of the single-head claim.}
We test models of 1B--7B parameters, and the effects are smaller in most of
those with 4B or more. Our procedure reports at most one head per model and does
not ablate sets of heads, so it cannot show how much of the function other heads
share or recover through self-repair \citep{mcgrath2023hydra}. We induce
departures by intervention and do not test whether naturally occurring language
confusion involves the same heads, and the five FLORES languages are all
European languages in Latin script.

\paragraph{Screening and measurement.}
Heads are selected by screening, so the results describe the selected heads,
not how common such heads are across models. The 125-prompt FLORES screen can
miss rare or prompt-specific effects, and selecting heads by their observed
effect favors upward noise; the seven FLORES heads still reach 0.077--0.517 on
the 2,375 prompts outside the screen, although these prompts also enter the
final selection. The decision to leave OLMo-3-7B's L14H25 out of the count was made after its LCB results were in, and the English-retention condition was written down after a first Qwen2.5-7B run (Appendix~\ref{app:head_screens}). The $\Delta$NLL and English-retention checks limit, but cannot
exclude, general damage to generation; language detection is imperfect despite
the detector checks; and effects are compared within a layer, not across
architectures.

\paragraph{Intervention semantics.}
Zeroing a head before the output projection can change downstream computation,
including other heads, and where the attention output is normalized after the
projection (Gemma-3, OLMo-2, OLMo-3) the norm rescales the change; holding it
fixed accounts for part of the effect in Gemma-3-1B. Mean ablation and
language-specific replacement insert fixed activations that may be atypical for
a given prompt or position, so neither isolates a semantic feature on its own.

\paragraph{Precision and checkpoints.}
Qwen2.5-1.5B runs in fp16 (fp32 in Section~\ref{sec:reply_language_mechanism}),
Qwen2.5-7B and OLMo-3-7B in bf16, and the other models in fp32. In bf16,
batched and single-prompt runs produce identical continuations on only 15 of 20
checked prompts in Qwen2.5-7B and 10 of 20 in OLMo-3-7B, so small effects in these two
models call for caution. In Qwen2.5-1.5B (fp16), two runs with the same settings give the same LPR but differently worded baseline replies for 184 of 2,189 prompts, so each change is paired with its own run's baseline. OLMo-2-1B
Base was screened in bf16, with its selected layer rerun in fp32, and its
FLORES continuations block the end-of-text token. Evaluations use greedy
decoding, with the repetition penalty set in Qwen2.5-Instruct's generation
configuration; sampling (temperature 0.7, two seeds) was checked only for Qwen2.5-1.5B on LCB, where L22H6's changes stay at $-0.273$ and $-0.274$ (monolingual) and $-0.245$ and $-0.276$ (crosslingual). Qwen2.5 models also receive their default English system prompt: in Qwen2.5-1.5B, L22H6's $C\to W$ is 0.500 with it, 0.379 with it translated into the prompt's language and 0.258 without a system prompt, and the head stays the strongest of the seven heads rerun in each setting; we did not run this check for the other Qwen2.5 models. The base and instruction-tuned
comparisons use released checkpoints rather than controlled training, and base
models get raw text while instruction-tuned ones get their chat template; the
Qwen2.5-1.5B format comparison separates the two only in part, so the
comparisons cannot attribute a change to any post-training stage.

\paragraph{Mechanism and steering.}
The mechanism checks cover two models and 480 crosslingual prompts, with 51
baseline-passing replies for Gemma-3-1B. The random-direction and second-language controls
separate what the name carries from the size of the change and from a bias toward
one replacement language, but they were added after the first value-patch results
and test two other languages per prompt language. The mask reproduces only some of
the failures from zeroing the head, so the head has other inputs. FLORES swap rates are measured on the prompts used to
estimate the language means, while LCB tests transfer to separate prompts.
Steering meets our improvement criterion only in the two Gemma models and does
not establish robustness in deployment, that response content is preserved,
or a general way to reduce language confusion.

\paragraph{Responsible use.}
The study uses publicly available models and benchmarks. Head-level
interventions can deliberately change a model's output language and may alter
response content, so practical use would require model-specific validation and
safeguards against unintended changes.

\bibliography{references}

\clearpage
\appendix

\section{Supplementary Protocol Details}
\label{app:protocol}

Each instruction-tuned input contains a single user turn, rendered with the
model's own chat template. Thinking is disabled where supported; templates that
insert a date use 2026-10-08. Qwen2.5 uses its default English system prompt,
and its generation configuration adds a repetition penalty (1.1 for 1.5B, 1.05
for 3B and 7B) to every greedy run. FLORES continuations use greedy decoding with
up to 40 new tokens, and LCB replies with up to 100. The FLORES screen uses 25
prompts per language: the five hand-written starts and the first 20 filtered
FLORES sentences. A model proceeds to the 2,500-prompt follow-up if at least one
head meets both the loss and English-retention conditions and reaches
$C\to W\geq 0.10$ on the screen (Section~\ref{sec:head_selection}). The FLORES
sentences are the first 495 aligned devtest sentences, in dataset order, for
which the versions in seven languages (the five prompt languages, Chinese and
Russian) all contain 20 to 300 characters and at least half of the characters
are letters. \texttt{langdetect} runs with its random seed fixed to zero, and a
continuation it cannot label counts as $\ne\ell_j$ (one continuation across the
seven heads found on FLORES).

For LCB, paired bootstrap confidence intervals compare an intervention and the
baseline on the same prompts; the percentile interval of 2,000 resamples is not
stratified by source. The benchmark's line-level scoring ignores lines shorter
than five words, counts a fastText label with confidence of 0.3 or less as
unknown, and averages pass rates over prompt sources, with the four languages
pooled within each source. $\Delta$LPR is computed over 778 to 800 of the 800
non-English monolingual and 1,159 to 1,190 of the 1,196 crosslingual prompts for
the heads in Table~\ref{tab:heads}. Chinese and
Japanese are segmented before the five-word filter in the 14-language extension, as in
the benchmark's scoring script \citep{marchisio-etal-2024-understanding}. Each run uses one NVIDIA H100 80GB GPU. We use fp16 for Qwen2.5-1.5B (fp32 in
Section~\ref{sec:reply_language_mechanism}), bf16 for OLMo-3-7B and Qwen2.5-7B,
and fp32 for the other models; each base checkpoint uses the precision of its
instruction-tuned model, except that the OLMo-2-1B base screen ran in bf16 with
its selected layer rerun in fp32. The OLMo-2-1B base run also blocks the
end-of-text token, which it otherwise emits right after most complete FLORES
sentences.

\section{FLORES Screens and Full-Set Follow-ups}
\label{app:head_screens}

Table~\ref{tab:full_flores_heads} summarizes the FLORES screens of the 12
evaluated models. The reported $C\to W$ and mean-ablation rates are from
the 2,500-prompt follow-up where one was run. The 0.10 threshold applies to the
125-prompt screen, so a full-set value can fall below it: OLMo-2-1B's L12H8
reaches 0.136 on the screen and 0.080 on the full set. Parentheses in the
$\Delta$NLL column give the mean change for the other heads in the selected
layer. The follow-up reruns every head in the layers containing the two heads
with the highest screened $C\to W$, whether or not they meet the primary conditions. It also reruns
the head with the highest screened $C\to W$ among those with $\Delta$NLL $\leq
0.1$ (all layers for Qwen2.5-1.5B), as an additional stricter low-loss comparison candidate. This cutoff is not the primary loss guard; primary eligibility uses $\Delta$NLL $\leq 1$ and at least 90\% English retention. Outside the rerun layers, no head that meets
both conditions exceeds a screened $C\to W$ of 0.064. Any English threshold from
0.6 to 0.95 selects the same head in each of the eight models with a follow-up.
This condition was written down after a first Qwen2.5-7B run selected
L0H25, whose removal keeps only 51\% of English continuations in
English; it changes none of the heads selected before. Lowering the
$\Delta$NLL bound to 0.25 changes no other selection, but below 0.72
Gemma-3-1B has no qualifying FLORES head (next: L5H0, 0.088 on the
screen).
On the 2,375 prompts not used in the screen, the seven heads' $C\to W$ is 0.077
to 0.517, close to 0.080 to 0.515 on all 2,500 and higher than on the 125 screen
prompts in four of the seven models. A dash marks a value not obtained because no FLORES head met the
screening rule; the heads found on LCB are in Table~\ref{tab:lcb_heads}.

\begin{table*}[t]
\centering
\small
\setlength{\tabcolsep}{3pt}
\begin{tabular*}{\linewidth}{@{\extracolsep{\fill}} l l c c l p{5.2cm}}
\toprule
Model & \shortstack{FLORES head\\($C\to W$)} & Mean & \shortstack{$\Delta$NLL\\(layer mean)} & Status & Note \\
\midrule
Qwen2.5-1.5B & L22H6 (0.500) & 0.450 & +0.228 (+0.004) & Follow-up & Base counterpart: 0.150 \\
Qwen2.5-3B & L27H13 (0.515) & 0.338 & +0.234 (+0.006) & Follow-up & Base counterpart: 0.202 \\
Qwen2.5-7B & L19H1 (0.200) & 0.018 & +0.026 (+0.001) & Follow-up & L0H25 (0.629) and L0H22 (0.466) fail English retention (0.512 and 0.518) \\
Qwen3-1.7B & L18H12 (0.324) & 0.028 & +0.128 (+0.003) & Follow-up & L0H3 (0.466) fails the loss condition ($\Delta$NLL $>1$) \\
Qwen3-4B & None (screen 0.024) & -- & -- & Screen only & LCB screened directly \\
Gemma-3-1B & L11H3 (0.413) & 0.089 & +0.720 ($-$0.012) & Follow-up & Same head has little effect in Base \\
Gemma-3-4B & L24H0 (0.214) & 0.012 & +0.202 ($-$0.005) & Follow-up & Base counterpart: 0.148 \\
OLMo-2-1B & L12H8 (0.080) & 0.056 & +0.136 (+0.022) & Follow-up & Small effect; also observed on LCB \\
OLMo-3-7B & L14H25 (0.134) & 0.007 & +0.202 (+0.000) & Follow-up & bf16; small LCB effect, no distinct head on LCB screen \\
Llama-3.2-1B & None (screen 0.048) & -- & -- & Screen only & LCB-only head: L8H25 \\
Llama-3.2-3B & None (screen 0.008) & -- & -- & Screen only & LCB-only head: L13H19 \\
SmolLM3-3B & None & -- & -- & Screen only & L1H12 reaches 0.880 but has $\Delta$NLL $+2.58$ and repeated \texttt{</think>} output \\
\bottomrule
\end{tabular*}
\caption{FLORES head screens and full-set follow-ups. Status: Follow-up, the
model went on to the 2,500 prompts; Screen only, no head met the screening rule,
so only the 125-prompt screen was run (screen values are the strongest result
there, not full-set estimates). Mean: $C\to W$ under pooled mean ablation.
OLMo-3-7B ran the follow-up but is not counted (see text).}
\label{tab:full_flores_heads}
\end{table*}

Seven heads found on FLORES are counted in the main analysis. OLMo-3-7B L14H25
is not: its LCB pass-rate changes are small ($-0.041$ on monolingual and
$-0.018$ on crosslingual prompts) and vanish under mean ablation ($+0.001$ and
$+0.015$; Table~\ref{tab:lcb_mean}), whereas every counted head keeps a drop with
its 95\% CI below zero on at least one LCB task. The separate LCB screen also
finds no comparably distinct head in OLMo-3-7B (largest drop 0.060). This
criterion was set after the LCB evaluation. Its small FLORES mean-ablation
effect (0.007) does not decide the question on its own, since mean ablation also
keeps only 6--22\% of the FLORES effect in Qwen2.5-7B, Qwen3-1.7B and the two
Gemma models (Appendix~\ref{app:zero_mean_diagnostics}).
Screening rates are not used to rank heads across model families, whose head
counts and normalization differ.

\section{LCB Results Across Models}
\label{app:lcb}

Table~\ref{tab:lcb_heads} gives the change in line-level pass rate under zero
ablation with its 95\% paired-bootstrap interval, next to the changes for the
same-layer control heads. Across models, the control heads change monolingual
LPR by $-0.011$ to $+0.010$ and crosslingual LPR by $-0.025$ to $+0.038$.
Gemma's crosslingual baseline LPR is already low (0.118 for 1B and 0.133 for 4B),
which limits the additional drop these models can show on that task.

\begin{table*}[t]
\centering
\small
\setlength{\tabcolsep}{3pt}
\begin{tabular*}{\linewidth}{@{\extracolsep{\fill}} l l c c c c}
\toprule
 & & \multicolumn{2}{c}{Selected head} & \multicolumn{2}{c}{Same-layer controls} \\
\cmidrule(lr){3-4}\cmidrule(lr){5-6}
Model & Head & $\Delta$ mono LPR & $\Delta$ cross LPR & $\Delta$ mono & $\Delta$ cross \\
\midrule
Qwen2.5-1.5B & L22H6  & $-$0.273 [$-$0.305, $-$0.240] & $-$0.276 [$-$0.303, $-$0.250] & $-$0.001 to $+$0.004 & $-$0.007 to $+$0.000 \\
Qwen2.5-3B   & L27H13 & $-$0.537 [$-$0.575, $-$0.503] & $-$0.449 [$-$0.478, $-$0.421] & $+$0.000 to $+$0.003 & $-$0.025 to $+$0.004 \\
Qwen2.5-7B   & L19H1  & $-$0.009 [$-$0.020, $+$0.003] & $-$0.074 [$-$0.089, $-$0.059] & $+$0.001 to $+$0.003 & $-$0.008 to $+$0.001 \\
Qwen3-1.7B   & L18H12 & $-$0.149 [$-$0.174, $-$0.123] & $-$0.508 [$-$0.535, $-$0.478] & $-$0.004 to $-$0.003 & $+$0.000 to $+$0.021 \\
Qwen3-4B     & L24H27 & $+$0.000 [$-$0.010, $+$0.010] & $-$0.076 [$-$0.093, $-$0.061] & $-$0.005 to $+$0.010 & $-$0.001 to $+$0.013 \\
Gemma-3-1B   & L11H3  & $-$0.628 [$-$0.662, $-$0.593] & $-$0.091 [$-$0.110, $-$0.073] & $-$0.005 to $+$0.003 & $-$0.023 to $+$0.038 \\
Gemma-3-4B   & L24H0  & $-$0.259 [$-$0.292, $-$0.227] & $-$0.107 [$-$0.126, $-$0.090] & $-$0.001 to $+$0.000 & $+$0.000 to $+$0.012 \\
OLMo-2-1B    & L12H8  & $-$0.272 [$-$0.302, $-$0.239] & $-$0.335 [$-$0.362, $-$0.308] & $+$0.000 to $+$0.004 & $-$0.003 to $+$0.004 \\
OLMo-3-7B    & L14H25 & $-$0.041 [$-$0.060, $-$0.024] & $-$0.018 [$-$0.033, $-$0.003] & $-$0.011 to $+$0.004 & $+$0.001 to $+$0.007 \\
Llama-3.2-1B & L8H25  & $-$0.016 [$-$0.026, $-$0.008] & $-$0.711 [$-$0.737, $-$0.684] & $-$0.009 & $+$0.001 \\
Llama-3.2-3B & L13H19 & $+$0.000 [$-$0.006, $+$0.006] & $-$0.335 [$-$0.363, $-$0.308] & $+$0.000 to $+$0.006 & $-$0.003 to $+$0.000 \\
\bottomrule
\end{tabular*}
\caption{LCB pass-rate change under zero ablation, with 95\% paired-bootstrap
intervals. Same-layer controls: range over three other heads of the layer (one,
L8H12, for Llama-3.2-1B). The Llama-3.2-1B, Llama-3.2-3B and Qwen3-4B heads
were found by screening LCB directly, the others on FLORES.}
\label{tab:lcb_heads}
\end{table*}

The direct LCB screen ablated every head on 100 crosslingual prompts. For the
three heads selected through this route, the unweighted share of replies scored
in both conditions that changed from pass to fail was 0.710 for Llama-3.2-1B
L8H25 (71/100), 0.323 for Llama-3.2-3B L13H19 (32/99), and 0.112 for Qwen3-4B
L24H27 (11/98). The largest such share for another head of the same layer was
0.040. These values count pass-to-fail
transitions rather than net changes in pass rate; in Qwen3-4B, 0.010 changed
from fail to pass, giving a net change of $-0.102$ on the screen. The
corresponding crosslingual changes on prompts outside the screen were $-0.711$,
$-0.336$, and $-0.073$ (95\% CI [$-0.090$, $-0.057$] for Qwen3-4B). SmolLM3-3B yielded no eligible head. These checks
distinguish the LCB-selected heads from the other heads in their layers and show
effects on prompts that were not used to select them.

Table~\ref{tab:lcb_mean} repeats the LCB evaluation with mean ablation. Each of
the ten counted heads keeps a drop with its 95\% CI below zero on the task it
affects most, 39--89\% of the zero-ablation drop, while OLMo-3-7B L14H25 does
not. The same table gives the two heads that the LCB screen selects instead of
the FLORES head (Section~\ref{sec:specificity}): Qwen2.5-1.5B L16H9 keeps its
crosslingual drop under mean ablation, and Qwen2.5-7B L0H1 does not.

On that screen, run on the eight models with a FLORES follow-up, the FLORES head
lowers the pass rate on the 100 screening prompts by 0.389 in Qwen2.5-3B, 0.485 in
Qwen3-1.7B, 0.343 in OLMo-2-1B, 0.660 in Gemma-3-1B and 0.340 in Gemma-3-4B (the
two Gemma models on monolingual prompts), the largest drop among the heads that
meet both conditions in each, against 0.052, 0.051, 0.081, 0.030 and 0.030 for the
next head. In Qwen2.5-1.5B, L22H6 lowers it by 0.220, second of 336 heads, behind
L16H9 (0.400); in Qwen2.5-7B, L19H1 lowers it by 0.040, tied for fourth of 780
heads that meet both conditions, behind L0H1 (0.232). OLMo-3-7B is described in
Appendix~\ref{app:head_screens}.

\begin{table*}[t]
\centering
\small
\setlength{\tabcolsep}{3pt}
\begin{tabular*}{\linewidth}{@{\extracolsep{\fill}} l l c c}
\toprule
Model & Head & $\Delta$ mono LPR & $\Delta$ cross LPR \\
\midrule
Qwen2.5-1.5B & L22H6  & $-$0.246 [$-$0.276, $-$0.215] & $-$0.247 [$-$0.273, $-$0.220] \\
Qwen2.5-3B   & L27H13 & $-$0.409 [$-$0.444, $-$0.373] & $-$0.331 [$-$0.358, $-$0.303] \\
Qwen2.5-7B   & L19H1  & $+$0.003 [$-$0.005, $+$0.010] & $-$0.035 [$-$0.047, $-$0.024] \\
Qwen3-1.7B   & L18H12 & $-$0.067 [$-$0.086, $-$0.048] & $-$0.230 [$-$0.257, $-$0.205] \\
Qwen3-4B     & L24H27 & $-$0.004 [$-$0.014, $+$0.006] & $-$0.040 [$-$0.052, $-$0.027] \\
Gemma-3-1B   & L11H3  & $-$0.244 [$-$0.275, $-$0.213] & $-$0.092 [$-$0.110, $-$0.075] \\
Gemma-3-4B   & L24H0  & $-$0.120 [$-$0.145, $-$0.095] & $-$0.053 [$-$0.066, $-$0.040] \\
OLMo-2-1B    & L12H8  & $-$0.128 [$-$0.152, $-$0.104] & $-$0.143 [$-$0.165, $-$0.123] \\
Llama-3.2-1B & L8H25  & $-$0.015 [$-$0.025, $-$0.005] & $-$0.357 [$-$0.384, $-$0.329] \\
Llama-3.2-3B & L13H19 & $-$0.005 [$-$0.011, $+$0.001] & $-$0.250 [$-$0.277, $-$0.225] \\
\midrule
OLMo-3-7B    & L14H25 & $+$0.001 [$-$0.011, $+$0.015] & $+$0.015 [$+$0.003, $+$0.028] \\
Qwen2.5-1.5B & L16H9  & $+$0.008 [$-$0.003, $+$0.018] & $-$0.277 [$-$0.303, $-$0.251] \\
Qwen2.5-7B   & L0H1   & $+$0.004 [$-$0.005, $+$0.013] & $-$0.005 [$-$0.013, $+$0.003] \\
\bottomrule
\end{tabular*}
\caption{LCB pass-rate change under mean ablation, with 95\% paired-bootstrap
intervals. Below the line: the uncounted OLMo-3-7B head and the two heads the
LCB screen selects instead of the FLORES head; under zero ablation, L16H9
changes the pass rates by $-0.009$ and $-0.459$, and L0H1 by $-0.116$ and
$-0.271$.}
\label{tab:lcb_mean}
\end{table*}

\subsection{Fourteen-language LCB extension}
\label{app:lcb14}

We extended LCB to its 14 non-English languages (Arabic, Chinese, French,
German, Hindi, Indonesian, Italian, Japanese, Korean, Portuguese, Russian,
Spanish, Turkish and Vietnamese) for six models: Qwen2.5-1.5B, Qwen2.5-3B,
Qwen3-1.7B, Gemma-3-1B, OLMo-2-1B, and Llama-3.2-1B. Under the
predefined criterion that a language's paired interval is below zero and its
change is below every same-layer control, the selected head lowers crosslingual
LPR in all 14 languages for five models and in 12 of 14 for Qwen2.5-1.5B.
Monolingual effects vary more: 10/14 for Qwen2.5-1.5B, 13/14 for Qwen2.5-3B,
12/14 for Qwen3-1.7B, 14/14 for Gemma-3-1B, 13/14 for OLMo-2-1B, and 6/14 for
Llama-3.2-1B. In Qwen2.5-3B, 273 of the 279 monolingual replies left
unscored without the head are written mainly in Han characters or kana, so the
monolingual changes for Korean, Arabic and Vietnamese are based on 9, 220
and 59 replies. The Qwen2.5-1.5B exceptions on crosslingual prompts are Japanese
and Russian; its language-specific pattern does not transfer uniformly to the
other models. These are per-language comparisons, and the intervals are not
adjusted for the number of languages.

\section{Detector and Content Checks}
\label{app:checks}

\paragraph{Alternative language detectors.}
We relabeled the full-set FLORES continuations with langid
\citep{lui-baldwin-2012-langid}, with fastText \citep{joulin2016fasttext}, and by
a vote that takes a label when at least two of langdetect
\citep{shuyo2010}, langid and fastText agree. Across all seven heads found on FLORES, $C\to W$
differs by at most 0.02 from the langdetect result. Under every detector, the
selected head ranks first among the heads rerun on 2,500 prompts, except in
Qwen3-1.7B, where L18H12 ranks second to L0H3, and Qwen2.5-7B, where L19H1
ranks third after L0H25 and L0H22. These three higher-ranked heads fail the
head-selection conditions.
For Qwen2.5-7B L19H1, $C\to W$ is 0.200 with langdetect and 0.200--0.208 with the
alternative detectors.

\paragraph{Content similarity after a switch.}
For prompts whose non-English continuation switches to English after ablation,
we measured cosine similarity between the prompt and the continuation using
Qwen3-Embedding-0.6B \citep{zhang2025qwen3embedding}. For scale, over all non-English FLORES prompts, not only
the switched ones, the next sentence of the same article has similarity 0.384 to
the prompt (0.370 when that sentence is taken in English), and a random sentence
0.179 (0.172 in English).

\begin{table}[t]
\centering
\small
\setlength{\tabcolsep}{3pt}
\begin{tabular*}{\linewidth}{@{\extracolsep{\fill}} l r r r r}
\toprule
Model & \shortstack{Switched\\prompts} & Before & After & \shortstack{Before/after\\reply} \\
\midrule
Qwen2.5-1.5B & 918 & 0.475 & 0.558 & -- \\
Qwen2.5-3B & 514 & 0.590 & 0.657 & 0.582 \\
Qwen3-1.7B & 739 & 0.635 & 0.683 & 0.600 \\
Gemma-3-1B & 873 & 0.639 & 0.559 & 0.586 \\
OLMo-2-1B & 101 & 0.702 & 0.668 & 0.607 \\
\bottomrule
\end{tabular*}
\caption{Content of the continuations that switch to English when the head is
zeroed. Switched prompts: their number, Before and After: cosine similarity
between the prompt and the continuation without and with the ablation,
Before/after reply: similarity between the two continuations, --: not measured.
Same-layer control heads switch at most two prompts each.}
\label{tab:content_similarity}
\end{table}

The departures do not always go to English. With the head zeroed, 918 of
Qwen2.5-1.5B's 1,250 departures go to English, but Qwen2.5-3B's 1,288 split into
English (514), Chinese (299) and Spanish (221), and 441 of Qwen2.5-7B's 500 go
to Chinese. In Qwen3-1.7B and Gemma-3-1B, the 739 and 873 continuations that
switch to English (Table~\ref{tab:content_similarity}) are most of the departures
($C\to W$ of 0.324 and 0.413 on 2,500 prompts).

In the five models, switched English continuations have a mean similarity of
0.558--0.683 to their prompts, above the reference values for the next sentence
of the same article (0.384 in the original language and 0.370 in English).
Because the references are computed on all non-English prompts rather than on
the switched ones, the comparison is only a rough indication of content
relatedness; it does not establish equivalent meaning or task correctness.

\section{Base and Post-training Checkpoints}
\label{app:checkpoint_details}

For the seven heads found on FLORES, the base-model values are in the Base column
of Table~\ref{tab:heads}; each head has a smaller full-set $C\to W$ in the
base model. Counting only the base continuations that stay in the prompt's
language without ablation changes the base-model values by at most 0.003. The OLMo-2-1B comparison follows L12H8 across the released base, SFT,
DPO, and instruction-tuned checkpoints \citep{olmo2024olmo2}, with FLORES $C\to W$ of 0.005, 0.028,
0.071, and 0.080. Base layer 12 is run in fp32, as the later checkpoints are.
Crosslingual LCB changes under zero ablation are $-0.247$, $-0.244$, and $-0.335$
for SFT, DPO, and the final checkpoint. The base checkpoint is tested without a
chat template and with the end-of-text token blocked. The final checkpoint adds
reinforcement learning with verifiable rewards after DPO, released through an
intermediate RLVR checkpoint, so the results do not isolate the effects of
individual training stages.

\section{Mechanism Check Details}
\label{app:mechanism}

\paragraph{Setup.}
The crosslingual prompts are drawn with a fixed seed from the LCB crosslingual
prompts that name the requested language in English, round robin over sources,
without using any model's replies. The 96 monolingual prompts for the phase
experiment are drawn from Qwen2.5-1.5B's baseline LCB rows with a fixed seed,
without filtering on earlier ablation results; the Italian and German prompts all
come from Okapi \citep{lai-etal-2023-okapi}, and the French and Spanish prompts
eight each from Dolly \citep{conover2023dolly}, the native set and Okapi. Both runs use fp32, eager attention, greedy decoding and
100 new tokens. A mask adds $-\infty$ to the attention logits of the selected head
only, at the masked key positions, in every decoding step after the first
generated token; the first generated token matches the clean run in every
condition. The nearby tokens are as many user-text tokens as the language name
has, the ones closest to it. The control head is the other head of the layer
with the largest mean attention from the last prompt token to the name in a
clean pass, chosen before any intervention. Attention is measured by teacher
forcing the clean reply. Over the first eight reply tokens of the 480 prompts,
L22H6 puts 0.375 of its attention on the name, against 0.018 for L22H7 and 0.001
for the nearby tokens; L11H3 puts 0.064, against 0.020 for L11H1, 0.079 for L11H2
and 0.005 for the nearby tokens. The exact McNemar tests \citep{mcnemar1947note} compare
the name mask with each control on the baseline-passing replies. The reading
fixed in advance requires the name mask to fail more replies than both controls
in a model, so the two tests per model form an intersection-union test
\citep{berger1982multiparameter} that needs no correction; Holm's correction
\citep{holm1979simple} across the two models gives $p=0.031$ for both.

\paragraph{Earlier selected-prompt check.}
An earlier Qwen2.5-1.5B L22H6 run used 96 crosslingual LCB prompts, including
Italian prompts selected from replies that had switched language under
full-head ablation. On this set, 95 replies passed at baseline, mean attention
to the requested-language name was 0.807, and masking that attention caused
seven baseline-passing replies to fail, against none for nearby tokens or for
the name at L22H8. All seven also failed when the head was zeroed in the
decoding steps. Because prompt selection depended in part on earlier
intervention outcomes, we report this run only as a supporting check.

\begin{table*}[t]
\centering
\small
\setlength{\tabcolsep}{4pt}
\begin{tabular}{@{}llrrrrrrrr@{}}
\toprule
 & & & \multicolumn{3}{c}{Masked} & & \multicolumn{2}{c}{Values replaced} & \\
\cmidrule(lr){4-6}\cmidrule(lr){8-9}
Model & Prompts & Pass & Name & Near & Ctrl & Zeroed & Same & Other & In other \\
\midrule
Qwen2.5-1.5B & 96      & 62  & 6  & 0 & 0 & 20  & 0 & 17 & 8/95 \\
             & 384 new & 261 & 28 & 0 & 0 & 85  & 1 & 73 & 36/378 \\
             & all 480 & 323 & 34 & 0 & 0 & 105 & 1 & 90 & 44/473 \\
\midrule
Gemma-3-1B   & 96      & 12  & 7  & 0 & 0 & 8   & 0 & 10 & 7/95 \\
             & 384 new & 39  & 25 & 0 & 2 & 28  & 0 & 33 & 17/376 \\
             & all 480 & 51  & 32 & 0 & 2 & 36  & 0 & 43 & 24/471 \\
\bottomrule
\end{tabular}
\caption{Mechanism checks on the 96 earlier and 384 new crosslingual LCB prompts.
Pass: replies that pass at baseline, of which the columns from Name to Other
count the ones that fail, Name, Near and Ctrl: the head's attention to the
language name or to as many nearby tokens masked, or the control head's attention
to the name, Zeroed: the head zeroed, Same and Other: the values the head reads
at the name replaced by their mean over prompts that name the same or another
language, In other: scored replies entirely in that other language, out of all
scored replies, all in the decoding steps after the first generated token. The masks on
the 96 prompts come from the earlier run, whose clean replies these runs
reproduce exactly.}
\label{tab:value_patch}
\end{table*}

\paragraph{Value patching and a larger sample.}
Two checks were written down, together with the code that runs them, before
they were run: the
same mechanism runs on 384 new crosslingual prompts, drawn by the same rule as
the 96 of Section~\ref{sec:reply_language_mechanism} and leaving those out, and a
value patch on all 480 prompts. Because the draw does not use model outputs, 31
of the 384 happen to be among the prompts of the selected-prompt check. The value patch
leaves the selected head's attention weights as computed and replaces the values
it reads at
the name positions: it adds $\sum_{j}\alpha_{tj}(\bar{v}-v_j)$ over those
positions to the head's attention output before the output projection, without
changing the shared key-value cache or directly intervening on other heads (the
layer has 2 key-value heads for 12 query heads in Qwen2.5-1.5B and 1 for 4 in
Gemma-3-1B).
$\bar{v}$ is the mean value at the name positions over the other prompts of the
run that name the same language, or over the prompts that name the other
language (German for French, Spanish and Italian prompts, French for German
ones). Patching each name position with its own value left every reply unchanged
in a smoke run. The prespecified readings are that the other-language patch fails
more baseline-passing replies than the same-language patch and puts more replies
entirely in the other language, both by exact McNemar tests on the 480 prompts,
unless the same-language patch already fails more replies than the nearby mask. A
note added after the smoke run and before the main runs counts the first reading
only together with the second, since the other-language patch changes the head's
output more. Both criteria are met in both models (Table~\ref{tab:value_patch}):
the other-language patch alone fails 89 and 43 replies that the same-language
patch keeps, with no reply the other way ($p\leq 2.3\times10^{-13}$), and alone
puts 44 and 24 replies in the other language, again with none the other way
($p\leq 1.2\times10^{-7}$), and the same-language patch does not fail significantly more
replies than the nearby mask (1 against 0 and 0 against 0). No reply is entirely
in the other language under the name mask, the nearby mask, the control head's
mask or zeroing the head (0 of 471 to 474 scored replies in each model). Few
replies change in the other direction: under the other-language
patch, 3 baseline-failing Qwen2.5-1.5B replies pass and none in Gemma-3-1B. In
Qwen2.5-1.5B, Italian replies again dominate: of 79 baseline-passing ones, the
name mask fails 20, zeroing the head 79 and the other-language patch 51.

\paragraph{Controls for the value patch.}
Because the other-language patch changes the head's output more, we added three
controls after the results above, with their readings written down before they
were run, together with the code that runs them. The random-direction patch moves
each name value $v_j$ by its distance to the other-language donor in a random
direction, $v_j+\lVert\bar v_{\mathrm{other}}-v_j\rVert\,u$, with one unit vector
$u$ per prompt shared by its name positions, so that the change to the head's
output is at least as large as under the other-language patch at each step; three
seeds. The second-language patch uses the donor of a second language (Spanish for
German prompts, Italian for Spanish and French prompts, French for Italian
prompts). In Gemma-3-1B, we also mask L11H2's attention to the name, the head that
attends to it most over the first eight reply tokens. All runs reproduce the
earlier clean replies on the 480 prompts. Every planned reading holds in both
models (Table~\ref{tab:value_controls}): the other-language patch fails more
replies than each random direction (87, 83 and 88 against 2, 1 and 2 discordant in
Qwen2.5-1.5B; 42, 40 and 37 against 0 in Gemma-3-1B; $p\leq 1.5\times10^{-11}$) and
puts more replies in the first donor language (44 and 24 against 0 for every
seed); the second-language patch puts more replies in its language than the
same-language patch (205 and 15 against 0; $p\leq 6.1\times10^{-5}$); and the
L11H3 name mask fails more replies than the L11H2 name mask (30 against 0;
$p=1.9\times10^{-9}$). Zeroing the head puts 8 Qwen2.5-1.5B replies entirely in
the second donor language and the name mask 4, all Italian prompts answered in
French, against none in Gemma-3-1B; no mask and no zeroing puts a reply entirely
in the first donor language. In a check added afterwards, the Qwen2.5-1.5B
results hold without the Italian prompts: the name mask fails 14 of the 244
baseline-passing replies, against none under the nearby mask, and the
other-language patch fails 39 and puts 22 replies entirely in the other
language, against 1 and 0 under the same-language patch and at most 5 failures
under each random direction ($p<10^{-3}$ for each comparison).

\begin{table*}[t]
\centering
\small
\setlength{\tabcolsep}{3pt}
\begin{tabular*}{0.75\textwidth}{@{\extracolsep{\fill}} l l r r r r}
\toprule
Model & Condition & Fail & First & Second & Change \\
\midrule
Qwen2.5-1.5B & Same language & 1 & 0 & 0 & 0.391 \\
(323 pass) & Other language & 90 & 44 & 2 & 2.007 \\
 & Random, seed 1 & 5 & 0 & 0 & 2.034 \\
 & Random, seed 2 & 8 & 0 & 0 & 2.028 \\
 & Random, seed 3 & 4 & 0 & 0 & 2.043 \\
 & Second language & 217 & 0 & 208 & 1.976 \\
\midrule
Gemma-3-1B & Same language & 0 & 0 & 0 & 0.102 \\
(51 pass) & Other language & 43 & 24 & 0 & 0.360 \\
 & Random, seed 1 & 1 & 0 & 0 & 0.367 \\
 & Random, seed 2 & 3 & 0 & 0 & 0.367 \\
 & Random, seed 3 & 6 & 0 & 0 & 0.368 \\
 & Second language & 25 & 0 & 16 & 0.335 \\
 & L11H2 name mask & 2 & 0 & 0 & -- \\
\bottomrule
\end{tabular*}
\caption{Value patch controls on the 480 crosslingual prompts. Fail:
baseline-passing replies that fail; First and Second: scored replies (471 to
476) entirely in the first donor language (German, or French for German prompts)
or the second (Spanish, Italian or French); Change: mean norm of the change to
the head's output per decoding step, not recorded for the mask (--).}
\label{tab:value_controls}
\end{table*}

\section{Zero and Mean Ablation Diagnostics}
\label{app:zero_mean_diagnostics}

\makeatletter
\setlength{\@fpsep}{8pt}
\setlength{\@fptop}{0pt}
\setlength{\@fpbot}{0pt plus 1fil}
\setlength{\@dblfpsep}{8pt}
\setlength{\@dblfptop}{0pt}
\setlength{\@dblfpbot}{0pt plus 1fil}
\makeatother

\begin{table*}[t]
\centering
\footnotesize
\setlength{\tabcolsep}{3pt}
\textit{(a) Head-output statistics and pooled-mean interventions}\\[3pt]
\begin{tabular*}{\textwidth}{@{\extracolsep{\fill}}llccccccc@{}}
\toprule
Model & Head & \shortstack{Norm\\rank} & \shortstack{Mean\\energy} & \shortstack{Lang.\\residual} & Zero & \shortstack{Prompt\\mean} & \shortstack{Cont.\\mean} & \shortstack{Minus\\cont. mean} \\
\midrule
Qwen2.5-1.5B & L22H6  & 1/12 & 0.18 & 0.62 & 0.501 & 0.451 & 0.431 & 0.027 \\
Qwen2.5-3B   & L27H13 & 2/16 & 0.38 & 0.71 & 0.515 & 0.338 & 0.372 & 0.059 \\
Qwen2.5-7B   & L19H1  & 2/28 & 0.38 & 0.77 & 0.202 & 0.021 & 0.008 & 0.070 \\
Qwen3-1.7B   & L18H12 & 1/16 & 0.34 & 0.56 & 0.324 & 0.028 & 0.155 & 0.138 \\
Gemma-3-1B   & L11H3  & 1/4  & 0.76 & 0.03 & 0.413 & 0.089 & 0.196 & 0.070 \\
Gemma-3-4B   & L24H0  & 2/8  & 0.56 & 0.73 & 0.214 & 0.012 & 0.011 & 0.003 \\
OLMo-2-1B    & L12H8  & 2/16 & 0.31 & 0.67 & 0.080 & 0.056 & 0.054 & 0.005 \\
\bottomrule
\end{tabular*}

\vspace{1.1em}
\textit{(b) Language-conditioned and control interventions}\\[3pt]
\begin{tabular*}{\textwidth}{@{\extracolsep{\fill}}lccccccccc@{}}
\toprule
 & & & & \multicolumn{3}{c}{Swapped-in language} & & & \\
\cmidrule(lr){5-7}
Model & \shortstack{Own-lang.\\mean} & \shortstack{English\\mean} & \shortstack{Other-lang.\\mean} & FLORES & \shortstack{LCB\\mono} & \shortstack{LCB\\cross} & \shortstack{Random\\same norm} & \shortstack{Half\\output} & \shortstack{Zero, norm\\fixed} \\
\midrule
Qwen2.5-1.5B & 0.052 & 0.486 & 0.530 & 0.182 & 0.043 & 0.019 & 0.544 & 0.071 & -- \\
Qwen2.5-3B   & 0.001 & 0.595 & 0.794 & 0.996 & 0.982 & 0.829 & 0.636 & 0.047 & -- \\
Qwen2.5-7B   & 0.000 & 0.188 & 0.352 & 0.417 & 0.122 & 0.128 & 0.228 & 0.017 & -- \\
Qwen3-1.7B   & 0.000 & 0.667 & 0.780 & 0.904 & 0.926 & 0.701 & 0.377 & 0.009 & -- \\
Gemma-3-1B   & 0.000 & 0.782 & 0.977 & 0.997 & 0.994 & 0.594 & 0.666 & 0.008 & 0.160 \\
Gemma-3-4B   & 0.000 & 0.222 & 0.796 & 0.989 & 0.985 & 0.137 & 0.291 & 0.000 & 0.234 \\
OLMo-2-1B    & 0.000 & 0.100 & 0.750 & 0.910 & 0.844 & 0.828 & 0.130 & 0.005 & 0.074 \\
\bottomrule
\end{tabular*}
\caption{Diagnostics for the seven heads found on FLORES, as $C\to W$ on 2,500
FLORES prompts except for the statistics in (a) and the Swapped-in language
columns in (b) (columns defined in the text). --: not run. Zero and Prompt mean
may differ slightly from Table~\ref{tab:heads} because these runs use
per-language batches. With any of three same-layer control heads swapped
instead, no LCB reply is in the swapped-in language.}
\label{tab:zero_mean_diagnostics}
\end{table*}

Mean replacement sometimes affects continuation language much less than zero
ablation (Table~\ref{tab:heads}). To examine this difference, we evaluated the
seven heads found on FLORES with additional interventions on 2,500 prompts.
These include pooled-mean and language-conditioned replacement, mean
subtraction, random-vector replacement, and output scaling. For models with
post-attention normalization, we also tested a control that holds the clean
normalization factor fixed. Table~\ref{tab:zero_mean_diagnostics} reports
$C\to W$ on the 2,500 FLORES prompts for each intervention. Panel (a) also gives
three statistics of the head's projected output $c_t$ over the user text and
continuation tokens: the rank of its mean norm in the layer (Norm rank), the share
$\lVert m\rVert^2/\mathbb{E}\lVert c_t\rVert^2$ of its energy in the mean vector
$m$ (Mean energy), and the share of the remaining variance explained by the
prompt's language, $\sum_\ell n_\ell\lVert m_\ell-m\rVert^2/\sum_t\lVert
c_t-m\rVert^2$ with $m_\ell$ the mean over the $n_\ell$ tokens of language $\ell$
(Lang.\ residual). Prompt mean and Cont.\ mean pool the head's output over prompt tokens
and over continuation tokens, and Minus cont.\ mean subtracts the latter. In
panel (b), the own-language, English and other-language means are the language
means of Section~\ref{sec:interventions}, with German for French, Spanish and
Italian prompts and French for German ones. Swapped-in language gives the share
of the 2,000 non-English FLORES continuations in that language, and of the scored
replies to non-English LCB prompts entirely in it. Random same norm replaces the
output with a random vector of the same norm, Half output scales it by 0.5, and
Zero, norm fixed holds the clean post-attention normalization factor fixed.

Replacing a head with the mean for the prompt's own language produces few
correct-to-wrong changes ($C\to W\leq 0.052$ in all seven models). Replacing it
with another language's mean yields higher $C\to W$ than zero ablation in every
model. The choice of replacement mean therefore matters: for Qwen3-1.7B,
Gemma-3-1B, and Gemma-3-4B, prompt-token pooled means yield $C\to W$ rates of
0.028, 0.089, and 0.012, respectively, whereas other-language means yield 0.780,
0.977, and 0.796. Continuation-token pooled means yield higher rates for
Qwen3-1.7B (0.155) and Gemma-3-1B (0.196), showing that the pooled-mean result
also depends on which tokens are averaged.

$C\to W$ alone is not specific to a language: a random vector of the same norm
moves about as many continuations as another language's mean in Qwen2.5-1.5B
(0.544 against 0.530), so the share in the swapped-in language is the measure of
redirection. The language means are estimated on the same FLORES prompts used
for the FLORES swap, whereas the LCB results test transfer to separate prompts.
In five models, at least 0.904 of continuations to non-English prompts under
other-language replacement are in the replacement language, and the FLORES means
put 0.844--0.994 of the monolingual LCB replies in that language. The shares are
lower for Qwen2.5-1.5B and Qwen2.5-7B. Holding the post-attention normalization
factor fixed reduces the zero-ablation rate for Gemma-3-1B from 0.413 to 0.160,
but changes the rate little for Gemma-3-4B (0.214 to 0.234) and OLMo-2-1B (0.080
to 0.074). A small effect under pooled-mean replacement therefore does not, by
itself, show that the head has little influence on continuation language.

\paragraph{Content of replies under replacement.}
For the crosslingual LCB replies that replacement with the requested language's
mean turns from fail to pass, we compute the cosine similarity
(Qwen3-Embedding-0.6B) to the baseline reply to the same prompt and to the
baseline reply to another prompt of the same task and language, with one fixed
pairing per prompt. The corrected replies (530 in Gemma-3-1B, 98 in Gemma-3-4B, 43 in
Qwen3-1.7B, 31 in Qwen2.5-3B, 20 in OLMo-2-1B, 16 in Qwen2.5-1.5B and 11 in
Qwen2.5-7B) have similarity 0.69--0.86 with the baseline reply to the same
prompt, against 0.15--0.21 with another prompt's for the replies whose pass
status changes. Under this replacement, no model's share of unscored replies or
of repetitive replies rises by more than 0.002. Similarity indicates overlapping
content, not the same meaning.

\section{Comparison with Translation Heads}
\label{app:mitra}

\citet{lasnier2026translation} identify \emph{language heads}, which select the
output language in few-shot translation, by activation patching on base models.
We compare our instruction-tuned heads with the per-head scores released by the
authors for Qwen3-1.7B-Base, Llama-3.2-1B, and Gemma-3-1B-pt
\citep[commit \texttt{daa721b}]{mitra2026repo}. \newpage
The scores cover 20 English-to
and English-from directions involving French, Spanish, Portuguese, Japanese,
Chinese, Swahili, Wolof, Hindi, Arabic, and Russian. We average each direction
over its examples, count how often our head is the direction's top head, and
rank its mean score across directions (Table~\ref{tab:translation_heads}).

\begin{table}[t]
\centering
\small
\setlength{\tabcolsep}{3pt}
\begin{tabular*}{\linewidth}{@{\extracolsep{\fill}} l l c c}
\toprule
Base model & Our instruct head & \shortstack{Directions\\topped} & \shortstack{Mean-score\\rank} \\
\midrule
Qwen3-1.7B & L18H12 & 20/20 & 1/448 \\
Llama-3.2-1B & L8H25 & 7/20 & 1/512 \\
Gemma-3-1B & L11H3 & 0/20 & 3/104 \\
\bottomrule
\end{tabular*}
\caption{Instruction-tuned heads under the Translation Heads ranking on released
base-model scores. In Llama-3.2-1B, L9H8 ties L8H25 in the number of directions
topped.}
\label{tab:translation_heads}
\end{table}

The match is strongest for Qwen3-1.7B and Llama-3.2-1B, but it does not show that
the heads serve the same function in base-model translation and
instruction-tuned language retention. Gemma-3-1B L11H3 ranks third by mean
score; another head, L15H2, tops all 20 directions.

\end{document}